\documentclass[sigconf,nonacm]{acmart}

\usepackage[T1]{fontenc}

\usepackage{microtype}

\usepackage{amsmath}
\usepackage{mathtools}
\usepackage{amsthm}
\usepackage{algorithm}
\usepackage{algpseudocode}
\usepackage{graphicx}
\usepackage{float}
\usepackage{xspace} 
\usepackage{booktabs}
\usepackage[most]{tcolorbox}
\usepackage{multicol}
\usepackage{enumitem}
\usepackage{colortbl}
\usepackage{seqsplit} %
\usepackage{booktabs}
\usepackage{multirow}
\usepackage{colortbl}
\usepackage{tabularx}
\usepackage{multirow}
\newcommand{\ourmethod}{{GRMP}\xspace}
\newcolumntype{C}{>{\centering\arraybackslash}X}

\newif\ifshowrevisions
\showrevisionsfalse
\ifshowrevisions
  \DeclareRobustCommand{\rev}[1]{\textcolor{red}{#1}}
  
  \newenvironment{revblock}{\color{red}}{}
\else
  \DeclareRobustCommand{\rev}[1]{#1}
  
  \newenvironment{revblock}{}{}
\fi
\DeclareRobustCommand{\chemmechkg}{ChemMechKG\xspace}
\DeclareRobustCommand{\chemmechqa}{ChemMechQA\xspace}

\usepackage{tabularx}  %
\usepackage{pifont} 
\usepackage{makecell} 

\newcommand{\cmark}{\textcolor{green!60!black}{\ding{51}}}
\newcommand{\xmark}{\textcolor{red}{\ding{55}}}

\title{Enhancing Multi-Step Chemical Mechanism Prediction with Graph-Guided Preference Data}
\providecommand{\correspondingauthor}{\authornote{Corresponding author.}}
\author{Yu Zhang}
\affiliation{%
  \institution{Shanghai Jiao Tong University}
  \city{Shanghai}
  \country{China}}
\email{cynthiazhang@sjtu.edu.cn}

\author{Ruijie Yu}
\affiliation{%
  \institution{Shanghai Jiao Tong University}
  \city{Shanghai}
  \country{China}}
\email{el\_iauk@sjtu.edu.cn}

\author{Xuelin Zhao}
\affiliation{%
  \institution{Shanghai Jiao Tong University}
  \city{Shanghai}
  \country{China}}
\email{xl\_zhao@sjtu.edu.cn}

\author{Jinming Yang}
\affiliation{%
  \institution{Shanghai Jiao Tong University}
  \city{Shanghai}
  \country{China}}
\email{yangjm67@sjtu.edu.cn}

\author{Feng Zhu}
\affiliation{%
  \institution{Shanghai Jiao Tong University}
  \city{Shanghai}
  \country{China}}
\email{fzchem@sjtu.edu.cn}

\author{Xiaokang Yang}
\affiliation{%
  \institution{Shanghai Jiao Tong University}
  \city{Shanghai}
  \country{China}}
\email{xkyang@sjtu.edu.cn}

\author{Yaohui Jin}
\affiliation{%
  \institution{Shanghai Jiao Tong University}
  \city{Shanghai}
  \country{China}}
\email{jinyh@sjtu.edu.cn}

\author{Yanyan Xu}
\correspondingauthor
\affiliation{%
  \institution{Shanghai Jiao Tong University}
  \city{Shanghai}
  \country{China}}
\email{yanyanxu@sjtu.edu.cn}

\renewcommand{\shortauthors}{Zhang et al.}

\begin{document}
\begin{abstract}
Predicting multi-step chemical reaction mechanisms requires identifying plausible intermediates and correctly ordering the elementary reaction steps. Existing computational methods focus on single-step elementary reaction prediction, limiting their ability to generalize to unseen reaction spaces. We introduce a graph-guided framework for multi-step chemical mechanism prediction that leverages a chemical mechanism knowledge graph (\chemmechkg) to guide large language models (LLMs). To improve pathway prediction, we develop a graph-aligned reaction mechanism predictor (GRMP) that uses graph neural networks to generate synthetic candidate pathways for pathway optimization. During inference, we design a relation retrieval (RR) module to extract relevant graph evidence from the KG for pathway refinement. Experiments show that \ourmethod achieves the highest pathway accuracy among all compared methods on \chemmechqa and oMe-Gold, reaching 48.1\% and 57.9\%, respectively. On a fixed sample of 110 out-of-distribution reactions from \chemmechqa{}-Hard, \ourmethod achieves 36.4\% top-1 pathway accuracy. LLM-as-Judge and human assessment further demonstrate its ability to generalize to unseen reactions. Code and data are available in the \href{https://anonymous.4open.science/r/CMKG-release/}{grmp}.
\end{abstract}

\maketitle

\begin{figure*}
    \centering
    \includegraphics[width=1\linewidth]{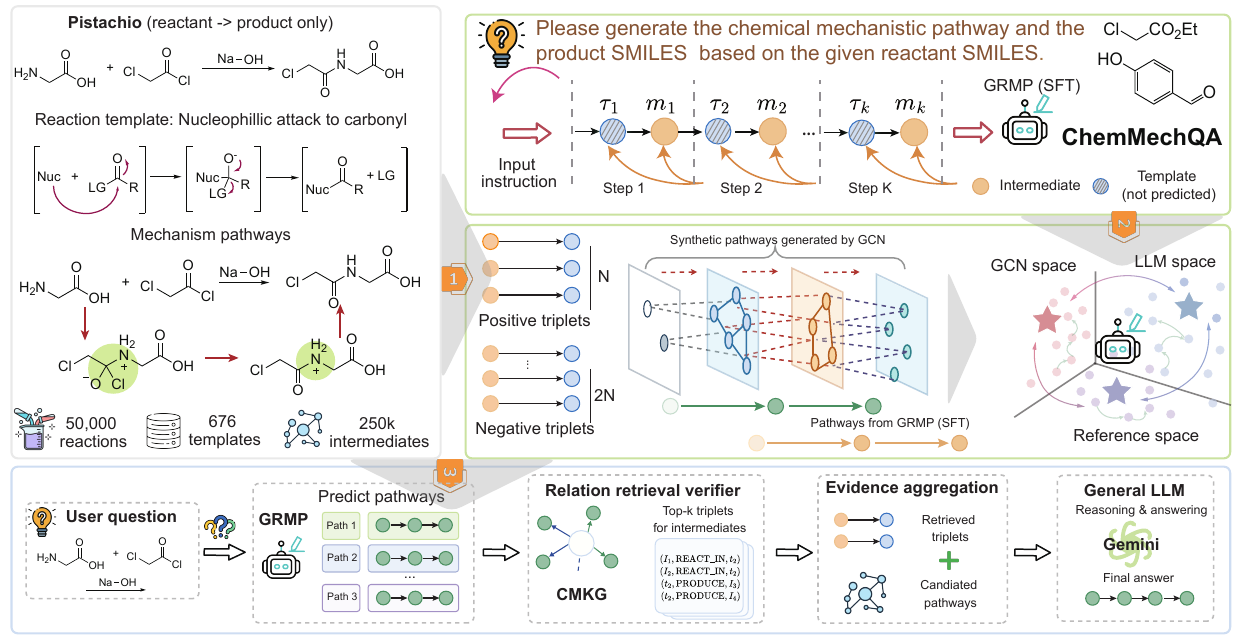}
    \caption{An illustration of our proposed framework. It contains \ourmethod, a graph-aligned LLM predictor, and a Relation Retrieval (RR) verifier for the mechanism pathway prediction and reasoning.}
    \label{fig:grmp}
\end{figure*}

\section{Introduction}\label{sec:introduction}

In organic chemistry, a reaction mechanism is a theoretical trajectory mapping electron redistribution during a chemical transformation. Chemists rely on reaction mechanisms as a foundational framework for predicting outcomes, typically represented using ``arrow pushing'' formalism~\cite{levy2017arrow}. By sequentially applying arrow-pushing logic to reactants, elementary intermediates are generated until final products are reached. However, due to the exponentially expanding search space of potential pathways, proposing accurate multi-step mechanistic routes remains a daunting challenge even for experienced domain experts.

Traditional methods for mechanism prediction involve three stages: (1) developing rule-based template systems to encode expert knowledge and label mechanistic data~\cite{bures2023organic, joung2024reproducing, chen2024large}; (2) retraining end-to-end machine learning models on mechanism data for single-step elementary reaction prediction~\citep{tavakoli2023ai}; (3) developing deep learning-based or generative methods to model electron redistribution to adhere to fundamental physical laws for single-step elementary reaction prediction~\cite{joung2025electron, chen2025predicting}. However, these methods are inherently constrained by their reliance on single-step elementary reaction data. Consequently, they are unable to  \emph{extrapolate across complex, multi-step mechanistic pathways}, even failing to \emph{generalize to unseen reactions} where the mechanistic logic deviates from known training patterns.

Recent advances in large language models (LLMs) have demonstrated remarkable reasoning capabilities across diverse chemical domains~\cite{zhang2025large, li2026collective, gottweis2026accelerating}. Nevertheless, current LLM-based applications in mechanism prediction remain predominantly confined to simplified multiple-choice selection tasks, leaving true multi-step mechanistic generation virtually unexplored~\cite{hao2026beyond}. This gap stems from a fundamental bottleneck: multi-step mechanistic prediction requires complex, long-chain scientific reasoning, where standard LLMs severely struggle with structural \emph{hallucinations} and a deficiency in chemical grounding~\cite{qiao2023reasoning, huang2023towards, huanglarge, wang2023knowledge}. To mitigate hallucinations in long-chain reasoning, Knowledge Graph (KG) integration has emerged as a promising paradigm to anchor LLMs with factual domain knowledge. However, existing KG-augmented approaches falter when directly deployed in complex chemical spaces. Extracting multi-hop mechanistic pathways from dense chemical graphs introduces substantial \emph{topological noise}. 

To address these issues, many studies have integrated Knowledge Graphs (KGs), which encapsulate factual information in a structured format, to improve the reasoning abilities of LLMs. However, existing KG-augmented approaches fail when directly deployed in complex chemical spaces. Extracting comprehensive, multi-hop mechanistic pathways from a chemical graph introduces severe topological noise. Conversely, shallow retrieval is insufficient for governing chemical steps, leaving downstream LLMs under-informed and still prone to cascading hallucinations.

To bridge this gap, we construct the Chemical Mechanism Knowledge Graph (\chemmechkg) by enriching reaction sequences with inferred intermediates derived from expert-validated elementary reaction templates. Building on \chemmechkg, we propose a graph-guided framework for multi-step chemical mechanism prediction (Figure~\ref{fig:grmp}). Specifically, we develop a GNN-based pathway generator to construct synthetic preference pairs for optimizing a graph-aligned reaction mechanism predictor (GRMP) via direct preference optimization (DPO). During inference, we design a relation retrieval (RR) module to extract relevant graph evidence from the KG for pathway refinement. Evaluations on \chemmechqa and oMe-Gold demonstrate that our framework achieves the highest pathway accuracy among the evaluated methods, while additional ablation studies investigate its generalization to held-out reaction classes and mechanism templates.

The main contributions of this work are as follows:

\begin{itemize}[topsep=2pt,itemsep=1pt,parsep=0pt,partopsep=0pt]
    \item We construct \chemmechkg, a knowledge graph that enriches raw reaction data with imputed intermediates using expert-curated templates.
   
    \item We propose a graph-guided framework for multi-step chemical mechanism prediction, in which GCN-generated candidate pathways provide synthetic preference data for pathway optimization. 
    
    \item We conduct extensive experiments and human evaluation, demonstrating that \ourmethod achieves state-of-the-art performance and generalizes effectively to unseen reactions.
   
\end{itemize}

\section{Related Work}\label{sec:related_work}

\textbf{Mechanism prediction.} Anticipating reaction mechanism inference at the elementary step level is crucial for predicting complex chemical outcomes. Early work focused on data curation by expert-annotated templates, such as MechFinder~\cite{chen2024large}, which subsequent works expanded to cover wider reaction distributions~\cite{chen2025predicting, joung2024reproducing}. Driven by these datasets, contemporary research applies deep learning to predict intermediate cascades~\cite{das2026deepmech}. Notably, graph-based models autoregressively generate consecutive intermediates over reaction graphs~\cite{joung2024reproducing}. FlowER~\cite{joung2025electron} models electron redistribution as a generative process for single-step predictions. LLMs are also applied to learn mechanistic reasoning for chemical reactions~\cite{dang2026learning,bran2503chemical}. Despite their success, these methods fail to autonomously extrapolate across multi-step mechanistic pathways and exhibit poor out-of-distribution (OOD) generalization when facing novel, unseen transformations.


\textbf{Synthetic data for LLM Training.} Synthetic data generation plays a pivotal role in expanding LLM capabilities beyond human-annotated corpora~\cite{kang2025demystifying}. Conventional selection strategies frequently filter synthetic data based on model confidence, prioritizing high-probability generations~\cite{huang2023large, huang2025self}. However, recent studies demonstrate that such a strategy yields diminishing marginal information gains and exacerbates model bias. Consequently, contemporary paradigms advocate for balanced curriculum mining, strategically incorporating both low-uncertainty (``easy'') and high-uncertainty (``hard'') samples to maximize training efficiency~\cite{setlur2024rl, zhang2025chemactor}. Nevertheless, systematically identifying informative samples remains non-trivial, particularly in complex scientific domains.

\section{Preliminary}\label{sec:preliminary}

\begin{figure*}[t]
    \centering
    \includegraphics[width=1\linewidth]{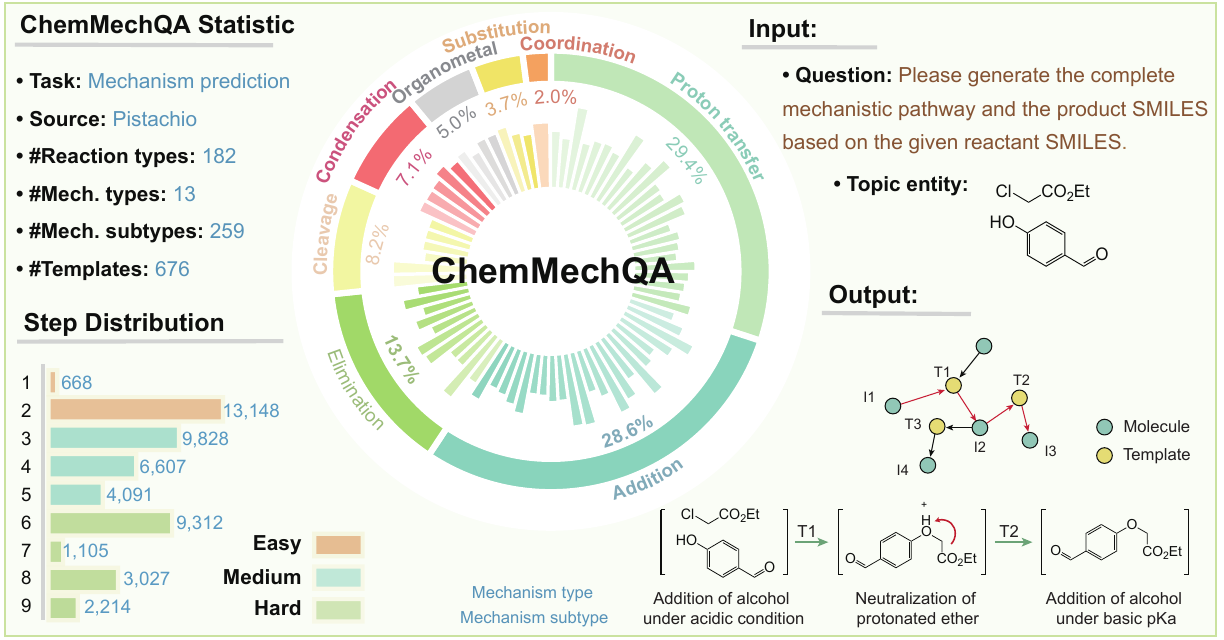}
    \caption{
    The distribution of our curated \chemmechqa.
    }
    \label{fig:cmqa_all}
\end{figure*}
\noindent\textbf{Chemical mechanism pathway}. 
A chemical reaction transforms reactants under given conditions into final products $\mathcal{P}$. We distinguish the molecular-state pathway $w=(m_0,\ldots,m_K)$ from its auxiliary elementary-template sequence $\boldsymbol{\tau}=(\tau_1,\ldots,\tau_K)$. The $i$-th elementary step is $m_{i-1}\xrightarrow{\tau_i}m_i$. Each $m_i$ is a molecular state represented by SMILES and may contain multiple dot-separated chemical components; $\tau_i$ is a reaction template represented by SMARTS.

In \chemmechkg, molecular states and templates are distinct entity types, $\mathcal{E}_{\mathrm{mol}}$ and $\mathcal{E}_{\mathrm{tpl}}$, with $\mathcal{E}=\mathcal{E}_{\mathrm{mol}}\cup\mathcal{E}_{\mathrm{tpl}}$. The corresponding annotated graph walk is $w_z=(m_0,\tau_1,m_1,\ldots,\tau_K,m_K)$. We use $e_j$ for a generic entity at graph position $j$, so that $e_{2i}=m_i$ and $e_{2i-1}=\tau_i$. Graph relations $r_j$ label the edges between these entities and are distinct from the template entities $\tau_i$. Thus, a pathway with $K$ elementary chemical steps has $2K$ graph hops and $2K+1$ entity positions. \ourmethod generates the molecular-state pathway $w$ during training and inference. 


\noindent\textbf{\chemmechqa}. We frame multi-step mechanism prediction as a specialized QA task over $\mathcal{G}$. Given a natural language query $q$, specifying the initial molecular state $m_0$, the objective is to learn a function $f$ that predicts a mechanistic pathway $w \in \mathcal{A}$ anchored in $\mathcal{G}$, such that $w = f(q, \mathcal{G})$. 

\section{Methods}
\subsection{\chemmechkg Curation} 
\label{sec:dataset-construction}
We build our mechanistic dataset on the processed Pistachio database~\cite{pistachio}, containing 15 million reactions. Building on the previous methodology for imputing mechanistic pathways~\cite{joung2024reproducing}, we derive structured mechanistic pathways from 50,000 experimentally reported reactions using 676 expert-annotated reaction templates proposed by~\cite{joung2025electron}. All mechanistic data are automatically reviewed and cleaned to enhance their consistency and reliability.

We define reaction templates and molecules as entities and labeled molecule--template connections as relations to construct \chemmechkg (Figure~\ref{fig:cmqa_all}). This yields 247,832 entities: 247,156 molecule nodes and 676 template nodes. Unlike datasets centered on isolated elementary steps, \chemmechkg provides pathway-level annotations of ordered intermediates, mechanism templates, natural-language descriptions, and reaction conditions in SMILES format. Details of our KG are in Table~\ref{tab:CMKG_statis} and Table~\ref{tab:cmqa_data_stat}. For model development, we extract the shortest reasoning paths, defined as sequences of consecutive triplets: $w_z = e_0 \xrightarrow{r_1} e_1 \xrightarrow{r_2} \dots \xrightarrow{r_{2K}} e_{2K}$, where $(e_{j-1}, r_j, e_j) \in \mathcal{G}$, $e_{2i}=m_i$, and $e_{2i-1}=\tau_i$ to construct \chemmechqa. Notably, all molecules and intermediates are excluded from the KG during model inference to avoid data leakage.

\begin{algorithm}[h]
\caption{\rev{Pair-Wise Synthetic Preference Data Construction}}
\label{alg:grmp}
\begin{revblock}
\small
\begin{algorithmic}[1]
\Require Annotated training set $\mathcal{D}_{\mathrm{train}}=\{(q_j,m_{0,j},w_j^*)\}_{j=1}^{n}$; graph $\mathcal{G}$
\Require Trained GCN predictor and SFT model $\pi_{\theta_s}$; graph-hop limit $T_{\max}$; $\alpha, \beta$
\Ensure Preference dataset $\mathcal{D}_{\mathrm{pref}}$ and companion reference SFT data
\State Initialize GCN-preferred set $\mathcal{D}_g$ and reference-preferred set $\mathcal{D}_r$ to $\varnothing$
\For{$(q_j,m_{0,j},w_j^*)\in\mathcal{D}_{\mathrm{train}}$}
    \State Generate $w_j^{\mathrm{sft}}$ from $\pi_{\theta_s}(\cdot\mid q_j,m_{0,j})$
    \State $s_s\gets R_I(w_j^{\mathrm{sft}},w_j^*)$
    \State Generate GCN molecular-state pathway $w_j^{\mathrm{gcn}}$ using $\mathcal{G}$ within $T_{\max}$ graph hops
    \If{$w_j^{\mathrm{gcn}}$ reaches the annotated product}
        \State Retain $w_j^{\mathrm{gcn}}$; set $s_g\gets R_I(w_j^{\mathrm{gcn}},w_j^*)$
    \Else
        \State Discard $w_j^{\mathrm{gcn}}$ as a preference candidate
    \EndIf
    \If{a retained GCN candidate exists \textbf{and} $s_g-s_s\geq\alpha$}
        \State Add $(q_j,w_j^{\mathrm{gcn}},w_j^{\mathrm{sft}})$ to $\mathcal{D}_g$
    \ElsIf{$s_s\leq\beta$}
        \State Add $(q_j,w_j^*,w_j^{\mathrm{sft}})$ to $\mathcal{D}_r$
    \EndIf
\EndFor
\State Approximately balance pair types by downsampling excess pairs in $\mathcal{D}_r$
\State Form $\mathcal{D}_{\mathrm{pref}}$ from $\mathcal{D}_g\cup\mathcal{D}_r$ by deduplicating $(q,w^+,w^-)$ triples and serializing for DPO
\State Save corresponding $w_j^*$ as companion SFT data, deduplicated by reaction
\State \Return $\mathcal{D}_{\mathrm{pref}}$ and the companion SFT data
\end{algorithmic}
\end{revblock}
\end{algorithm}
\subsection{GCN for Pathway Prediction}
We employ a query-dependent Graph Neural Network (GNN) model to learn the topological dependencies within \chemmechkg and predict the mechanism pathway. Our implementation uses a two-layer graph convolutional network (GCN)~\cite{kipf2016semi} encoder with a multilayer perceptron (MLP) decoder. Based on the definitions in the preliminary, we map each entity node $e \in \mathcal{E}$ to an initial textual embedding $\mathbf{z}_e \in \mathbb{R}^d$ using a pre-trained text embedding model~\cite{chen2024bge}. 

We designate the reactant nodes in the input query $q$ as query-relevant nodes, denoted $\mathcal{E}_q \subseteq \mathcal{E}$. The query-conditioned initial node representation $\mathbf{h}_e^0$ is defined as:
\begin{equation}
\mathbf{h}_e^0 = \texttt{Init}\left(\mathbf{z}_e,\; \mathbb{1}[e \in \mathcal{E}_q] \cdot \mathbf{z}_q\right), \quad \forall e \in \mathcal{E}.
\label{eq:init}
\end{equation}
where $\mathbf{z}_q \in \mathbb{R}^d$ is the query embedding derived from the input reactant SMILES, $\mathbb{1}[\cdot]$ denotes the indicator function, and $\texttt{Init}$ is a single-layer MLP that fuses the intrinsic node embedding with the query vector for query-relevant nodes, while leaving non-query nodes unchanged.
We then perform $L$ message-passing layers over the entity graph. At each layer $l \in \{1, \dots, L\}$, the embedding of node $e$ is updated by aggregating features from its localized neighborhood:
\begin{equation}
\mathbf{h}_e^{l} = \texttt{Update}\Big( \mathbf{h}_e^{l-1},\; \texttt{Agg}\big( \mathcal{M}_{e}^{l} \big) \Big),
\label{eq:message-passing}
\end{equation}

where $\mathcal{M}_{e}^{l} = \left\{ \texttt{Msg}\left(\mathbf{h}_e^{l-1}, \mathbf{h}_{e'}^{l-1}\right) \mid \exists r:\ (e',r,e) \in \mathcal{G} \right\}$ represents the set of incoming messages from neighboring entities. The message function $\texttt{Msg}$ parameterizes the interaction between connected entities as an element-wise product:
\begin{equation}
\texttt{Msg}(\mathbf{h}_e, \mathbf{h}_{e'}) = \mathbf{h}_e \odot \mathbf{h}_{e'},
\end{equation}
where $\odot$ denotes element-wise multiplication. The $\texttt{Agg}$ function aggregates neighbor messages, and $\texttt{Update}$ non-linearly combines them with the previous node representation.

We construct pathways by iteratively selecting and applying reaction templates. The two-layer MLP decoder assigns an edge score $s(m_t,\tau;q)$ to the current molecular state $m_t$ and each candidate template $\tau$, using the query-conditioned GCN representations. During training, the encoder and decoder are optimized with a binary edge-classification loss that distinguishes positive from negative edges. During inference, we normalize the candidate template scores:
\begin{equation}
p(\tau\mid m_t,q)=
\frac{\exp s(m_t,\tau;q)}
{\sum_{\tau'\in\mathcal{E}_{\mathrm{tpl}}}\exp s(m_t,\tau';q)}.
\end{equation}
We select the highest-scoring template and apply its SMARTS transformation to $m_t$ to generate the next molecular state $m_{t+1}$. Repeating template selection and application constructs the pathway until termination or the predefined maximum graph-hop threshold $T_{\max}$ is reached. The resulting alternating molecule--template sequence is denoted by $w_z^{\text{gcn}}$. Its molecular-state subsequence provides candidate responses for the preference construction described below.

\subsection{Graph-Aligned LLM for Mechanism Prediction (GRMP)} 


Given a chemical reasoning prompt $q$ and an initial molecular state $m_0$ identified within $q$, \ourmethod predicts an ordered molecular-state pathway $w=(m_0,\ldots,m_K)$, as defined in the preliminary. Let $w^*$ denote the molecular-state subsequence of the shortest annotated walk in \chemmechkg. Reaction templates provide graph information but are not included in the LLM output. The workf

\textbf{Supervised fine-tuning (SFT) stage.} We first apply SFT~\cite{ouyang2022training} to learn these annotated pathways. Let $y^*=\operatorname{Serialize}(w^*)$ be its reference token sequence. For $N$ target tokens, the objective is:
\begin{equation}
    \mathcal{L}_{\text{SFT}}(\theta_s) = -\sum_{t=1}^{N}\log P_{\theta_s}(y_t^*\mid y_{<t}^*,q,m_0),
\end{equation}
where $\theta_s$ denotes the SFT model parameters.

\textbf{Graph-grounded synthetic data for preference optimization.} For each training reaction, we extract all molecular-state subsequences of the SFT model and GCN-derived pathways to ensure they share the same format. GCN-derived pathways provide synthetic and diverse candidate pathways for constructing pair-wise preference data.

Below, $w^{\mathrm{sft}}$ and $w^{\mathrm{gcn}}$ denote these SFT and GCN molecular-state pathways, respectively. We extract intermediates from each candidate and the reference, canonicalize them, and compute their multiset matching rates against the reference. Let $I(w)$ be the multiset of canonicalized intermediates in $w$, and $c_y(u)$ the multiplicity of intermediate $u$. The intermediate matching rate is reference-normalized multiset recall:
\begin{equation}
R_I(w,w^*)=
\frac{\sum_u\min\{c_y(u),c_{w^*}(u)\}}
{|I(w^*)|}.
\label{eq:path-sim}
\end{equation}
We define a criterion for preference selection termed the intermediate recall: $s_s=R_I(w^{\mathrm{sft}},w^*)$ and $s_g=R_I(w^{\mathrm{gcn}},w^*)$. This score measures intermediate coverage without encoding step order.

Based on this criterion, we construct two types of preferred--rejected pairs using different thresholds $\alpha$ and $\beta$. The workflow of synthetic data selection is in Algorithm~\ref{alg:grmp}. A \emph{GCN-preferred} pair $(w^{\mathrm{gcn}},w^{\mathrm{sft}})$ is formed when $s_g-s_s\geq\alpha$. A \emph{reference-preferred} pair $(w^*,w^{\mathrm{sft}})$ is formed when $s_s\leq\beta$. If a reaction satisfies both conditions, we retain only its GCN-preferred pair. We downsample excess reference-preferred pairs to approximately balance the two types, then deduplicate by the $(q,w^+,w^-)$ triple, where $w^+$ and $w^-$ are the preferred and rejected responses Finally, we use the resulting pairs to construct $\mathcal{D}_{\mathrm{pref}}$ for DPO training.

\textbf{Preference Optimization.} Starting from the SFT model, we optimize the target policy $\pi_\theta$ using DPO~\cite{rafailov2023direct}, with the frozen SFT model as $\pi_{\mathrm{ref}}$:
\begin{equation}\small
\begin{split}
\mathcal{L}_{\mathrm{DPO}}(\theta)
= &-\mathbb{E}_{(q,w^+,w^-)\sim\mathcal{D}_{\mathrm{pref}}}
\Bigg[\log\sigma\\
&\Bigg(\lambda\log\frac{\pi_\theta(w^+\mid q,m_0)}{\pi_{\mathrm{ref}}(w^+\mid q,m_0)}\\
&-\lambda\log\frac{\pi_\theta(w^-\mid q,m_0)}{\pi_{\mathrm{ref}}(w^-\mid q,m_0)}
\Bigg)\Bigg],
\end{split}
\end{equation}
Here, $\pi(w\mid q,m_0)$ denotes the probability of the serialized molecular-state pathway. The function $\sigma$ is the sigmoid function and $\lambda$ scales the preference term. The objective favors reference pathways or GCN alternatives selected for greater intermediate coverage over the SFT candidates.

\subsection{Relation Retrieval for Inductive Reasoning}
During inference, the RR module extracts factual triplets as structural priors for inductive reasoning. Given an input reaction query $q$, GRMP generates a candidate molecular-state pathway $w=(m_0,\ldots, \\ m_K)$. For each state $m_i$, entity linking selects a molecular-state node $a_i\in\mathcal{E}_{\mathrm{mol}}$ using Tanimoto similarity:
\begin{equation}
a_i = \arg\max_{e \in \mathcal{E}_{\mathrm{mol}}} \operatorname{Tanimoto}\!\left(
\operatorname{FP}(m_i),
\operatorname{FP}(e)
\right).
\end{equation}
The linked nodes form the anchor set $E_{\text{anchor}}=\{a_i\}_{i=0}^{K}$. For each anchor $a_i$, we retrieve the relation triplets in its $h$-hop graph neighborhood $\mathcal{N}_h(a_i)$. Here, $h$ counts graph edges rather than elementary chemical steps; the retrieved neighborhoods can include both molecular-state and template nodes. Finally, the retrieved triplets guide the general-purpose LLM to generate the final pathway $w_{\text{final}}$, using the prompt in Appendix: 
\begin{equation} \small
    w_{\text{final}} = \text{LLM}_{\text{general}}\Big(\text{Prompt}(\bigcup_{a_i\in E_{\text{anchor}}}\mathcal{N}_h(a_i)\big)\Big)
\end{equation}

\noindent\textbf{Leakage-controlled retrieval.}
Before retrieval, we remove the molecule nodes and incident edges associated with each test reaction from \chemmechkg, preventing RR from directly accessing its annotated pathway.

\begin{table*}[t]
\centering
\definecolor{ablationblue}{HTML}{0072B2}
\caption{Overall performance on \chemmechqa and oMe-Gold. For general-purpose LLMs, results are averaged over five runs, each using a randomly sampled subset of 200 samples from ChemMechQA. $\text{GRMP}^*$ excludes final general-purpose LLM refinement. Bold and underline indicate the best and second-best performance for each metric.}
\label{tab:baseline}
\definecolor{resultblue}{HTML}{0072B2}
\setlength{\tabcolsep}{4.5pt}
\renewcommand{\arraystretch}{1.2}
\fontsize{9pt}{10pt}\selectfont

\begin{tabularx}{\textwidth}{>{\raggedright\arraybackslash}X*{8}{c}}
\toprule
\multirow{2}[2]{*}{\textbf{Method}} & \multicolumn{4}{c}{\textbf{\chemmechqa $\uparrow$}} & \multicolumn{4}{c}{\textbf{oMe-Gold $\uparrow$}} \\
\cmidrule(lr){2-5} \cmidrule(lr){6-9}
& Validity (\%) & MPA (\%) & I-hit (\%) & PA (\%) & Validity (\%) & MPA (\%) & I-hit (\%) & PA (\%) \\
\midrule
\rowcolor[HTML]{f0f0f0}
\multicolumn{9}{l}{\textit{\textbf{General LLMs for mechanism reasoning}}} \\

GPT-5.2 ~\cite{singh2025openai}
& 70.3$_{\pm 0.09}$ & 10.0$_{\pm 0.09}$ & 8.1$_{\pm 0.51}$ & 5.0$_{\pm 0.22}$ & 90.1 & 23.1 & 29.1 & 23.6 \\

GPT-5.2 + Few-shot ~\cite{wei2022chain}
& 78.9$_{\pm 0.85}$ & 13.1$_{\pm 0.27}$ & 13.0$_{\pm 0.00}$ & 0.0$_{\pm 0.35}$
& 90.6 & 17.8 & 11.3 & 18.3 \\
Gemini-3.1-Pro ~\cite{team2023gemini}
& 90.1$_{\pm 0.30}$ & 52.2$_{\pm 0.20}$ & 7.3$_{\pm 0.31}$ & 10.0$_{\pm 0.27}$ & 90.5 & \underline{57.8} & 37.9 & 34.9 \\
DeepSeek-V4-Pro~\cite{deepseekai2026deepseekv4} & 95.0$_{\pm 0.24}$ & 48.0$_{\pm 0.30}$ & 12.9$_{\pm 1.01}$ & 5.0$_{\pm 0.40}$  & 87.6 & 45.3         & 25.0 & 22.5 \\

Llama-3.1-8B~\cite{grattafiori2024llama}
& 40.0$_{\pm 0.77}$ & 0.0$_{\pm 0.00}$ & 0.81$_{\pm 0.00}$ & 0.0$_{\pm 0.00}$
& 73.5 & 12.6 & 3.3 & 2.7 \\
Qwen2.5-7B ~\cite{hui2024qwen2} & 55.5$_{\pm 0.50}$ & 1.0$_{\pm 0.00}$ & 1.0$_{\pm 0.22}$ & 0.0$_{\pm 0.00}$ & 58.4 & 14.9 & 4.7 & 3.2 \\
ChemDFM ~\cite{zhao2025developing}       & 44.2 & 0.0  & 0.0  & 0.0  & 17.4 & 0.0          & 0.0  & 0.0  \\
\midrule
\rowcolor[HTML]{f0f0f0}
\multicolumn{9}{l}{\textit{\textbf{Mechanistic models for multi-step mechanism prediction}}} \\
\rowcolor{white} 
Reactron~\cite{chen2025predicting}       & 93.0 & \textbf{68.6} & 29.3 & 11.8 & 91.8 & 8.6  & 33.6 & 5.2  \\
Graph2SMILES~\cite{joung2024reproducing} & \underline{96.7} & 61.9 & 37.6 & 33.5 & \underline{93.9} & 7.0 & 28.3 & 5.2  \\
FlowER ~\cite{joung2025electron}         & \textbf{100.0} & 51.8 & 31.4 & 27.2 & -- & -- & -- & -- \\
\midrule
\rowcolor[HTML]{f0f0f0}
\multicolumn{9}{l}{\textit{\textbf{Graph-aligned framework}}} \\
$\text{GRMP}^*$ (w/o general LLMs)        & 92.1 & 48.3 & \underline{50.3} & \underline{36.9} & 83.4 & 53.6 & \underline{61.3} & \underline{41.8} \\
\rowcolor{ablationblue!10}
\textbf{\ourmethod (Gemini-3.1-Pro)}    & 93.9 & \underline{67.8} & \textbf{64.3} & \textbf{48.1} & 89.3 & \textbf{73.8} & \textbf{63.6} & \textbf{57.9} \\
\bottomrule
\end{tabularx}
\end{table*}
\section{Experiments}
Our experiments aim to address the following research questions
(RQs):
\begin{enumerate}
    \item \textbf{RQ1}: How does \ourmethod compare with the general-purpose LLMs and prediction baselines? 
    \item \textbf{RQ2}: How do the key components of \ourmethod contribute to generalization on unseen reactions?
    \item \textbf{RQ3}: How sensitive is performance to preference
thresholds, node embeddings, and pathway length?
    \item \textbf{RQ4}: How do automated and human assessments characterize pathway plausibility? 
\end{enumerate}

\subsection{Experimental Setup}
\noindent\textbf{Datasets.} We evaluate on \chemmechqa, our primary dataset constructed directly from \chemmechkg, oMe-Gold (196 literature-verified pathways)~\cite{xu2025omebench}, and \chemmechqa{}-Hard. \chemmechqa{}-Hard comprises 2,009 reactions drawn from eleven held-out reaction classes and 24 unseen mechanism templates, providing an out-of-distribution (OOD) evaluation setting. We conduct an ablation study and human evaluation using a fixed 110-reaction sample from \chemmechqa{}-Hard. Dataset statistics appear in Appendix~\ref{app:data_cons}.

\noindent\textbf{Baselines and metrics.} We compare nine LLM configurations and three mechanistic baselines in three categories: 1) LLM reasoning methods, 2) mechanistic models including Reactron~\cite{chen2025predicting}, FlowER~\cite{joung2025electron}, and Graph2SMILES~\cite{joung2024reproducing}, and 3) our graph-aligned method. Following prior work~\cite{joung2024reproducing,joung2025electron}, we report Validity, Main Product Accuracy (MPA), Intermediate Hit (I-hit), Pathway Accuracy (PA), and Sequence Rank (SR). PA measures exact pathway matching. MPA is evaluated from the final responses of model generation. Full metric definitions appear in Appendix~\ref{app:implementation}.

\noindent\textbf{Implementation.} We utilize Qwen2.5-7B-Instruct as the foundation model. For LLM training in both the SFT and DPO stages, we adopt LoRA-based fine-tuning. The batch size for SFT is set to 4, and the learning rate is set to 5e-6. To construct synthetic data for preference optimization, we design a two-layer GCN~\cite{kipf2016semi} with BGE embeddings~\cite{chen2024bge} that follows an encoder--decoder architecture. We use 2$\times$ A6000-48G GPUs for training. During inference, our framework includes Gemini-3.1-Pro for final refinement. Gemini generates separate candidate sets for each $k$, so top-$k$ scores need not be monotonic. Experimental settings and hyperparameters are detailed in Appendix~\ref{app:implementation}.

\subsection{Main Results (RQ1)}
We evaluate the mechanism prediction performance of \ourmethod against state-of-the-art baselines across two benchmark datasets: \chemmechqa and oMe-Gold. 
Table~\ref{tab:baseline} reports top-1 results. Notably, $\text{GRMP}^*$ represents our proposed graph-aligned LLM utilizing our two-stage training but without inductive reasoning. \ourmethod (Gemini-3.1-Pro) achieves the highest reported Pathway Accuracy (PA) among the evaluated configurations on both datasets: 48.1\% on \chemmechqa and 57.9\% on oMe-Gold. Besides, our method outperforms the strongest mechanistic baseline (Graph2SMILES, 33.5\% PA) by 14.6 percentage points on \chemmechqa. While general-purpose LLMs like Gemini-3.1-Pro and DeepSeek-V4-Pro exhibit high SMILES validity (up to 95.0\%), their full-pathway accuracy remains poor, frequently falling at or below 10.0\%. Therefore, high molecular validity does not imply accurate reconstruction of complete pathways. On the independent, literature-verified oMe-Gold benchmark, Graph2SMILES and Reactron each achieve 5.2\% PA. These results indicate limited generalization capabilities for the two evaluated mechanistic models on this benchmark. Instead, our method achieves 57.9\% PA on oMe-Gold, exceeding Gemini-3.1-Pro (34.9\%) by 23.0 percentage points and GPT-5.2 + CoT (36.9\%) by 21.0 percentage points. Further, $\text{GRMP}^*$ without final LLM refinement also achieves 41.8\% PA. 

Next, we stratify \chemmechqa by the number of elementary steps $L$: Easy ($L\leq2$), Medium ($2<L\leq5$), and Hard ($5<L\leq9$). Table~\ref{tab:ablation_hops} shows that Graph2SMILES leads in Easy PA (62.1\%), but its PA falls to 11.2\% on Hard reactions. On Hard reactions, $\text{GRMP}^*$ achieves 29.0\% PA and 67.1\% SR; refinement increases these to 32.0\% and 73.1\%. However, we notice that the gains are metric-dependent: refinement slightly reduces PA from 42.0\% to 41.6\% on Medium reactions. Overall, these results highlight the advantages of $\text{GRMP}^*$ over conventional baselines on complex, multi-step reactions, while indicating that the benefits of refinement vary across difficulty levels and evaluation metrics. 

\begin{table}[ht]
\centering
\definecolor{ablationblue}{HTML}{0072B2}
\caption{Performance comparison across different reaction complexities on \chemmechqa. Bold and underline indicate the best and second-best performance for each difficulty level.}
\label{tab:ablation_hops}
\definecolor{resultblue}{HTML}{0072B2}
\setlength{\tabcolsep}{2.5pt} 
\renewcommand{\arraystretch}{1.15}
\fontsize{9pt}{10pt}\selectfont
\begin{tabularx}{\linewidth}{l*{4}{C}}
\toprule
\textbf{Method} & \textbf{MPA (\%)$\uparrow$} & \textbf{I-hit (\%)$\uparrow$} & \textbf{PA (\%)$\uparrow$} & \textbf{SR (\%)$\uparrow$} \\
\midrule

\rowcolor[HTML]{f0f0f0}
\multicolumn{5}{l}{\textit{\textbf{Easy Complexity}}} \\
FlowER       & 63.6 & 57.0 & 53.5 & 60.5 \\
Graph2SMILES & \textbf{73.4} & \underline{64.2} & \textbf{62.1} & \underline{67.4} \\
$\text{GRMP}^*$ & 64.2 & 40.9 & 32.4 & 62.3 \\
\rowcolor{ablationblue!10}
\textbf{\ourmethod} & \underline{70.8} & \textbf{71.4} & \underline{57.2} & \textbf{67.5} \\

\midrule
\rowcolor[HTML]{f0f0f0}
\multicolumn{5}{l}{\textit{\textbf{Medium Complexity}}} \\
FlowER       & 54.0 & 31.2 & 24.5 & 38.9 \\
Graph2SMILES & \underline{65.0} & 38.6 & 31.6 & 46.0 \\
$\text{GRMP}^*$ & 55.7 & \underline{60.4} & \textbf{42.0} & \underline{81.6} \\
\rowcolor{ablationblue!10}
\textbf{\ourmethod} & \textbf{67.3} & \textbf{68.9} & \underline{41.6} & \textbf{86.0} \\

\midrule
\rowcolor[HTML]{f0f0f0}
\multicolumn{5}{l}{\textit{\textbf{Hard Complexity}}} \\
FlowER       & \underline{38.8} & 23.3 & 7.8  & 35.1 \\
Graph2SMILES & \textbf{48.1} & 28.1 & 11.2 & 43.1 \\
$\text{GRMP}^*$ & 37.1 & \textbf{41.9} & \underline{29.0} & \underline{67.1} \\
\rowcolor{ablationblue!10}
\textbf{\ourmethod} & 38.1 & \underline{41.1} & \textbf{32.0} & \textbf{73.1} \\

\bottomrule
\end{tabularx}
\end{table} 
\subsection{Ablation Study (RQ2)}
As reported in Table~\ref{tab:ablation_dpo}, we evaluate four variants, including base predictors, LLM refinement, DPO parameters, and RR configurations on the fixed 110-reaction samples from \chemmechqa{}-Hard.  We make the following observations:
 
\textbf{Foundational models.} Qwen2.5-7B consistently outperforms Llama-3.1-8B under the setting of \textit{w/o} LLM refinement, achieving higher top-1 I-hit (19.9\% vs. 11.3\%), and MPA (14.5\% vs. 11.8\%). This superiority persists when paired with Gemini-3.1-Pro (30.9\% vs. 21.8\% top-1 MPA). These comparisons support Qwen2.5-7B as the
base predictor among the two evaluated backbones.

\textbf{LLM refinement.} Integrating LLM refinement improves MPA in the reported comparisons, but its effect on PA depends on the base model and refiner itself. For the SFT-only predictor, integrating Gemini-3.1-Pro with Qwen2.5-7B boosts MPA from 14.5\% to 30.9\% at top-1, while PA remains 12.7\%. For the $\text{GRMP}^*$ with Gemini-3.1-Pro, it substantially enhances top-1 MPA from 35.4\% to 49.1\% and top-1 PA from 27.2\% to 34.5\%.


\begin{table*}[t]
\centering
\definecolor{ablationblue}{HTML}{0072B2}
\definecolor{ablationteal}{HTML}{009E73}
\setlength{\tabcolsep}{4.5pt}
\renewcommand{\arraystretch}{1.22}
\fontsize{9}{10}\selectfont
\caption{Ablation study on the components of \ourmethod using fixed 110-reaction samples from \chemmechqa{}-Hard, where the best results are in bold.
}
\begin{tabularx}{\textwidth}{>{\raggedright\arraybackslash}X*{12}{r}}
\toprule
\textbf{Configuration}
& \multicolumn{3}{c}{\textbf{Validity (\%) $\uparrow$}}
& \multicolumn{3}{c}{\textbf{I-hit (\%) $\uparrow$}}
& \multicolumn{3}{c}{\textbf{MPA (\%) $\uparrow$}}
& \multicolumn{3}{c}{\textbf{PA (\%) $\uparrow$}} \\
\cmidrule(lr){2-4}\cmidrule(lr){5-7}\cmidrule(lr){8-10}\cmidrule(l){11-13}

\textbf{(Components of methods)}
& Top-1 & Top-3 & Top-5
& Top-1 & Top-3 & Top-5
& Top-1 & Top-3 & Top-5
& Top-1 & Top-3 & Top-5 \\

\midrule
\rowcolor{gray!9}\multicolumn{13}{l}{\textit{\textbf{Mechanistic models}}} \\
\rowcolor{white}
FlowER & \textbf{100.0} & \textbf{100.0} & \textbf{100.0} & 18.6 & 19.0 & 20.2 & 14.6 & 14.6 & 14.6 & 10.9 & 10.9 & 10.9 \\
Graph2SMILES & 89.1 & 90.9 & 90.9 & 15.7 & 17.7 & 18.2 & 12.7 & 14.6 & 14.6 & 7.3 & 9.1 & 9.1 \\
\midrule
\rowcolor{gray!9}\multicolumn{13}{l}{\textit{\textbf{Ablation of general LLMs (SFT-only predictors; w/o DPO, w/o RR)}} } \\
\rowcolor{white}
\multicolumn{13}{l}{\textit{SFT-only predictor: Llama-3.1-8B}} \\
 \textit{w/o} LLM refinement & 66.7 & 70.0 & 72.7 & 11.3 & 12.5 & 13.2 & 11.8 & 12.7 & 13.6 & 8.2 & 9.1 & 10.0 \\
+ GPT-5.2 & 88.4 & 90.0 & 90.9 & 23.0 & 25.0 & 25.8 & 14.5 & 16.4 & 18.2 & 6.4 & 7.3 & 8.2 \\
+ Gemini-3.1-Pro & 90.0 & 92.7 & 93.6 & 25.4 & 29.2 & 30.5 & 21.8 & 24.5 & 26.4 & 10.0 & 10.9 & 11.8 \\
\addlinespace[2pt]\multicolumn{13}{l}{\textit{SFT-only predictor: Qwen2.5-7B}} \\
\textit{w/o} LLM refinement & 85.4 & 90.6 & 94.5 & 19.9 & 20.7 & 21.0 & 14.5 & 20.0 & 21.0 & 12.7 & 12.7 & 13.6 \\
+ GPT-5.2 & 88.0 & 91.8 & 92.7 & 25.8 & 28.5 & 30.1 & 24.5 & 26.3 & 28.2 & 7.3 & 10.9 & 12.7 \\
+ Gemini-3.1-Pro & 90.3 & 93.6 & 94.5 & 28.4 & 28.2 & 29.5 & 30.9 & 33.6 & 33.6 & 12.7 & 12.7 & 13.6\\
\midrule
\rowcolor{gray!9}\multicolumn{13}{l}{\textit{\textbf{Ablation of training components (Qwen2.5-7B; w/o RR)}}  } \\
w/ DPO (reference-only) & 78.2  & 80.0 & 85.4 & 14.1 & 24.6 & 28.2 & 3.6 & 7.3 & 9.1 & 5.5 & 12.7 & 12.7 \\
\rowcolor{gray!5}
w/ DPO ($\alpha=0.3,\,\beta=0.3$) & 93.6 & 93.6 & 96.3 & 23.0 & 25.8 & 29.4 & 35.4 & 36.3 & 38.2 & 27.2 & 36.3 & 36.3 \\
\rowcolor{gray!9}\multicolumn{13}{l}{\textit{ \textbf{Ablation of relation retrieval components (Qwen2.5-7B + DPO; w/ Gemini-3.1-Pro)}}} \\

$\text{GRMP}^*$ & \underline{96.4} & \underline{96.4} & \underline{96.4} & 29.8 & 38.3 & 38.4 & 49.1 & 52.7 & 52.7 & 34.5 & 34.5 & 34.5 \\
\ourmethod (\textit{w/} RR, template-hidden KG) & 90.9 & 93.6 & 93.6 & \underline{34.7} & \underline{40.3} & \underline{40.7} & \underline{66.4} & \underline{70.9} & \underline{75.5} & \textbf{37.3} & \underline{38.1} & \textbf{40.9} \\
\rowcolor{ablationblue!10}
\ourmethod \textbf{(\textit{w/} RR, standard KG)} & 89.3 & 96.3 & 96.3 & \textbf{34.8} & \textbf{40.9} & \textbf{46.6} & \textbf{66.4} & \textbf{72.7} & \textbf{76.3} & \underline{36.4} & \textbf{38.2} & \underline{38.2} \\
\bottomrule
\end{tabularx}

\label{tab:ablation_dpo}
\end{table*}

\textbf{DPO provides the largest sequential gain for predicting mechanism pathways.} To isolate the contribution of graph-guided preference construction, we compare our approach with a reference-only DPO baseline that constructs preference pairs directly from reference pathways and incorrect SFT predictions. The results show that reference-only DPO achieves a top-1 PA of only 5.5\%, compared with 27.2\% for graph-guided DPO, an absolute improvement of 21.7 percentage points. Next, we discuss the contribution of DPO for the prediction performance. Compared with \ourmethod (SFT-only) without final LLM refinement, DPO with $\alpha=\beta=0.3$ raises top-1 MPA from 14.5\% to 35.4\% and PA from 12.7\% to 27.2\%, respectively. Collectively, using GCN-derived synthetic preference pairs for DPO training effectively improves OOD generalization, providing a structurally aligned foundation for subsequent LLM verification.



\textbf{The RR module primarily improves intermediate and product prediction}. For each prediction query, the standard RR refers to retrieving a KG that excludes its own reference pathway. Template-hidden KG additionally excludes the 24 held-out template nodes and their related edges. Compared to the $\text{GRMP}^*$ (\textit{w/o} RR), RR with template-hidden KG raises top-1 I-hit from 29.8\% to 34.7\% and MPA from 49.1\% to 66.4\%, respectively. Compared to $\text{GRMP}^*$ (\textit{w/o} RR), top-1 PA increases from 34.5\% to 37.3\% with template-hidden RR and to 36.4\% with standard RR. These consistent gains confirm that incorporating external mechanistic knowledge via relation retrieval provides critical structural priors, effectively constraining search spaces and guiding the LLM toward globally coherent reaction pathways.


\begin{figure}[ht]
    \centering
    \includegraphics[width=\columnwidth]{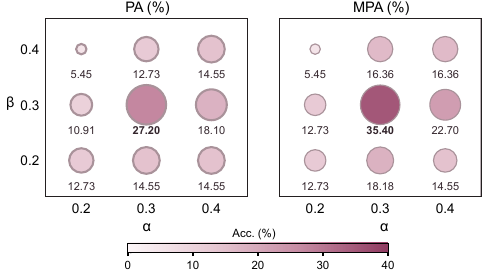}
    \caption{Top-1 PA and MPA (\%) for nine preference-threshold configurations on the 110-reaction OOD samples. Both metrics are highest at $\alpha=\beta=0.3$ among the evaluated settings.}
    \label{fig:sensitivity-v1}
\end{figure}

\subsection{Sensitivity and Failure Analyses (RQ3)}
Based on the ablation study results, we conduct sensitivity analysis of the DPO threshold parameters ($\alpha, \beta \in [0.2, 0.4]$) and GCN embeddings, respectively.

\textbf{Preference thresholds.} As we discussed before, increasing $\alpha$ selects fewer GCN-preferred pairs with larger intermediate-recall gains over SFT predictions, whereas increasing $\beta$ broadens the candidate pool for reference-preferred pairs. Therefore, we design nine configurations where $\alpha, \beta \in [0.2, 0.4]$ for sensitivity analysis. In Figure~\ref{fig:sensitivity-v1}, the highest observed PA and MPA occur at $\alpha=\beta=0.3$ (27.2\% and 35.4\%). At $(0.2,0.4)$, both scores fall to 5.45\%. Increasing $\alpha$ to 0.4 with $\beta=0.3$ yields 18.1\% PA and 22.7\% MPA. These results favor moderate selection thresholds, with both more permissive and more restrictive settings yielding lower PA and MPA than $\alpha=\beta=0.3$ in the evaluated configurations.

\begin{figure}[h]
\centering 
\includegraphics[width=\columnwidth]{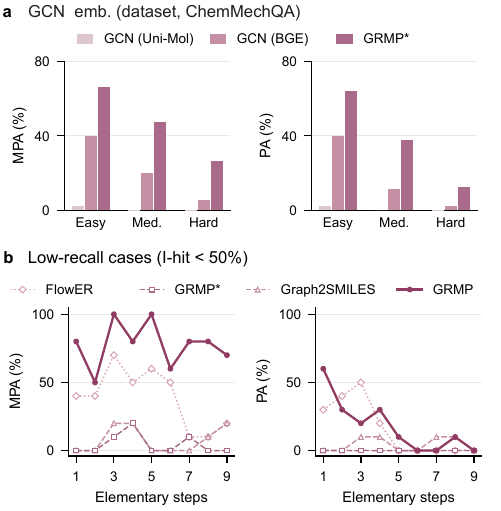}
\caption{GCN embedding comparison and failure-case analysis on \chemmechqa. (a) The performance of GCN with different embeddings and our GRMP$^*$ across different reaction complexities; (b) Failure-case analysis and optimization results of our $\text{GRMP}^*$.}
\label{fig:sensitive-analysis-v2}
\end{figure}

\textbf{GCN embeddings.} We evaluate the effectiveness of embeddings used for 
the GCN predictor on \chemmechqa. Note that we do not evaluate it on OOD reactions because standard GCN models exhibit severely degraded performance across all embedding variants in OOD regimes. Figure~\ref{fig:sensitive-analysis-v2}a compares 3D molecular embeddings (Uni-Mol) against textual semantic embeddings (BGE) alongside our complete $\text{GRMP}^*$ across reaction complexities (Easy, Medium, Hard). Text-based BGE embeddings consistently outperform Uni-Mol across all complexities for the isolated GCN module (e.g., 39.8\% vs. 2.1\% MPA on Easy, 11.4\% vs. 0.0\% MPA on Medium, and 2.0\% vs. 0.0\% MPA on Hard). Building upon BGE's superior semantic representation, our integrated $\text{GRMP}^*$ achieves substantial reasoning gains, reaching 64.2\%, 37.7\%, and 12.2\% PA on Easy, Medium, and Hard tasks, respectively. Although full PA naturally decreases as reaction complexity grows due to step-wise error propagation, BGE's strong domain alignment makes it the optimal structural foundation for our predictor.

\textbf{Failure cases analysis across pathway lengths.} Defining failure cases as samples with I-hit below 50\% in the initial $\text{GRMP}^*$ predictions, we select a 90-subset from \chemmechqa to investigate model robustness. Figure~\ref{fig:sensitive-analysis-v2}b analyzes MPA and PA metrics across elementary steps, comparing baseline methods and our proposed framework \ourmethod. For the unrefined predictor $\text{GRMP}^*$, performance remains consistently low across all step counts. However, incorporating LLM refinement (\ourmethod) dramatically improves these failure cases across all steps. \ourmethod achieves perfect MPA (100.0\%) at Step 3 and Step 5 and maintains a notable PA lead over both G2S and $\text{GRMP}^*$ on short-to-medium sequences (e.g., 60.0\% PA at Step 1 and 30.0\% at Step 2). While PA declines on long-chain reactions ($\ge 6$ steps), post-hoc refinement effectively prevents catastrophic error propagation, preserving robustness across complex multi-step mechanisms.

\subsection{LLM-as-Judge and Human Evaluation (RQ4)}
To evaluate the plausibility and chemical rigor of our generated pathways, we conduct an \textbf{LLM-as-Judge} evaluation alongside \textbf{human assessment} on the 110 sampled OOD reactions. Specifically, we prompt GPT-5.2 to evaluate the pathways across five expert-defined criteria detailed in Appendix. We evaluate three methods: FlowER, the reference pathway (Gold), and our \ourmethod. 

\begin{figure}[ht]
\centering 
\includegraphics[width=0.8\columnwidth]{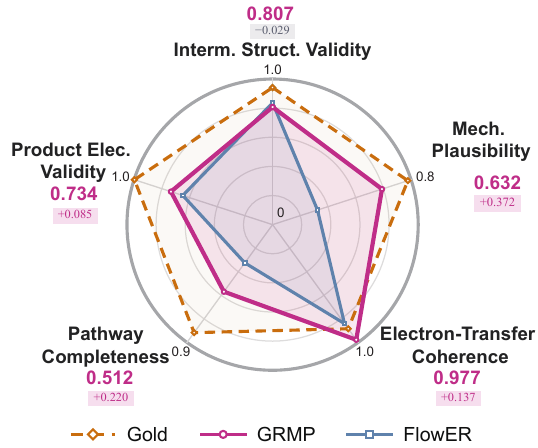}
\caption{LLM-as-Judge assessments of reference pathways (Gold), FlowER, and GRMP based on five chemical criteria.}
\label{fig:llm-as-judge-result}
\end{figure}

Figure~\ref{fig:llm-as-judge-result} shows that \ourmethod achieves higher scores than FlowER on four of the five criteria. Notably, \ourmethod obtains 0.632 on \textbf{mechanistic plausibility} and 0.512 on \textbf{pathway completeness}, representing absolute improvements of 0.372 and 0.220 over FlowER, respectively. Overall, after being verified by a powerful general-purpose LLM, \ourmethod produces more chemically plausible and complete reaction pathways than FlowER, highlighting its effectiveness in modeling multi-step reaction mechanisms.

Next, we invite three organic chemistry Ph.D. candidates to independently evaluate the predicted pathways and score the results based on the same criteria. We then systematically compute the average scores across eleven distinct reaction classes, indexed from 1 to 11 as mapped in Table~\ref{tab:CMKG_space}. Statistical analysis reveals that \ourmethod achieves an average score exceeding 0.60 in five out of the eleven reaction categories. Further, we supplement three representative case studies in Appendix~\ref{app:cases}. These include: 
(1) a typical failure case; (2) a reaction example where experts judge the model's prediction to be more reasonable than the reference label (Figure~\ref{fig:case_study1_2}); and (3) an example where both the reference and the prediction are logical under the LLM judge, yet human experts prefer the model's route on chemical grounds (Figure~\ref{fig:case_study3}). These evaluations provide complementary evidence that the framework can generate plausible alternatives to a single reference annotation, while experimental validation would be required before any such route could be treated as a discovered mechanism.

\section{Conclusion}
We introduce a novel approach for reliable multi-step reaction mechanism prediction by integrating structural chemical knowledge graphs with the generative power of large language models. We construct \chemmechkg using expert-annotated reaction templates and intermediates. We propose a graph-aligned framework consisting of GRMP and an RR verifier for multi-step mechanism pathway prediction. Empirical experiments and human evaluation demonstrate the effectiveness of our approach, showcasing its potential to generate plausible mechanistic pathways.


\bibliographystyle{ACM-Reference-Format}
\bibliography{main}

\appendix

\raggedbottom
\renewcommand{\topfraction}{0.95}
\renewcommand{\bottomfraction}{0.95}
\renewcommand{\textfraction}{0.05}
\renewcommand{\dbltopfraction}{0.95}
\setcounter{topnumber}{4}
\setcounter{totalnumber}{6}
\begin{table*}[t]
\caption{Comparison of representative mechanistic resources. \chemmechqa contains 259 categories of mechanistic subtypes (13 parent types). ``MI?'', ``NL'', ``MP'', and ``Cond'' indicate the presence of mechanistic intermediates, natural language descriptions, mechanistic pathways, and reaction conditions in SMILES format, respectively. }
\centering

\begingroup
\setlength{\tabcolsep}{4pt}
\renewcommand{\arraystretch}{1.15}
\small

\begin{tabular}{lccccccc}
\toprule
\textbf{Dataset} & 
\textbf{\#Reactions} & 
\makecell{\textbf{\#Reaction } \\
\textbf{Types}} & 
\makecell{\textbf{\#Mech. types/}\\\textbf{subtypes}} &
\textbf{MI?} & 
\textbf{NL?} &
\textbf{MP?} & 
\textbf{Cond?} \\
\midrule 
\citet{tavakoli2023rmechdb} & 5k+ & -- & 7 & \cmark & \xmark & \xmark & \xmark \\
\citet{chen2024large}  & 31k & 86 & 13 & \xmark & \xmark & \cmark & \xmark\\
\citet{joung2024reproducing}  & 192k & 86 & 13 &  \cmark & \xmark & \xmark & \cmark\\
\citet{hao2026beyond}  & 275 & 86 & 13 &  \xmark & \cmark & \xmark & \xmark\\
\citet{xu2025omebench} & 2.8k & -- & 8 & \cmark & \cmark & \cmark & \xmark\\
\chemmechqa & 50k & 182 & 259  & \cmark & \cmark & \cmark & \cmark\\
\bottomrule
\end{tabular}

\endgroup
\label{tab:dataset_comparison}
\end{table*}
\section{Chemical Mechanism Knowledge Graph (\chemmechkg) Construction} 
\label{app:data_cons}
 \chemmechqa provides template-imputed pathways for multi-step mechanism prediction. Table~\ref{tab:dataset_comparison} compares the source corpus with representative mechanistic resources, which differ in annotation granularity and coverage. 

To construct the mechanistic dataset, we leverage the Pistachio patent database (2025Q2 release), which comprises more than 15 million organic reactions. To ensure data quality, we implement a rigorous preprocessing pipeline. First, we canonicalize all reaction components (reactants, products, agents, solvents, and catalysts) in SMILES format. Second, building upon the methodology introduced by~\cite{joung2024reproducing}, we impute mechanistic pathways from experimentally reported reactions. After filtering samples with invalid mechanistic pathways and eliminating duplicates, we obtain 942,343 reactions and 8,222,189 elementary reactions with mechanistic pathways across 182 reaction classes and 676 reaction templates (see Table~\ref{tab:CMKG_statis}).

Finally, from this refined pool, we randomly partition 40,000, 5,000, and 5,000 reactions for training, validation, and testing, respectively (Table~\ref{tab:cmqa_data_stat}). For OOD evaluation, we isolate 11 reaction classes and 24 mechanism templates absent from training to construct \chemmechqa{}-Hard, which contains 2,009 reactions (Table~\ref{tab:CMKG_space}). Together, the resulting resource comprises 52,009 reactions and 217,441 elementary reaction steps.

\begin{table}[ht]
\centering
\caption{Key statistics of \chemmechkg.}
\label{tab:CMKG_statis}

\setlength{\tabcolsep}{4pt}
\renewcommand{\arraystretch}{1.12}
\small

\begin{tabular}{@{}lr@{}}
\toprule
\textbf{Statistic} & \textbf{Count} \\
\midrule
Entities                & 247,832 \\
Molecule nodes          & 247,156 \\
Template nodes          & 676 \\
Triples                 & 422,160 \\
Reaction types          & 182 \\
Mechanism types         & 13 \\
Mechanism subtypes      & 259 \\
\midrule
\multicolumn{2}{@{}l}{\textit{KG analysis}} \\
\midrule
Maximum hops            & 18 \\
Average hops            & 8.4432 \\
Max. template in-degree & 10,220 \\
Min. template in-degree & 1 \\
\midrule
$2 \leq \text{hop} < 6$   & 13,816 \\
$6 \leq \text{hop} < 12$  & 20,526 \\
$12 \leq \text{hop} \leq 18$ & 15,658 \\
\bottomrule
\end{tabular}
\end{table}

 \begin{table}[ht]
\centering
\caption{Split statistics for \chemmechqa and oMeBench.
The oMeBench test split corresponds to oMe-Gold, with 722
elementary steps after preprocessing.
$^*$ denotes the class- and template-held-out OOD split.}
\label{tab:cmqa_data_stat}

\setlength{\tabcolsep}{2pt}
\renewcommand{\arraystretch}{1.12}
\small

\begin{tabular}{@{}l l r r r r@{}}
\toprule
Dataset & Split & Mol. & Rxns. & Steps & Types \\
\midrule
\multirow{2}{*}{oMeBench}
& Train & 12,684 & 2,508 & 10,619 & -- \\
& Test  & 1,151  & 196   & 722    & -- \\
\midrule
\multirow{3}{*}{\chemmechqa}
& Train & 197,759 & 40,000 & 168,545 & 167 \\

& Val.  & 25,162  & 5,000  & 21,354  & 132 \\
& Test  & 24,981  & 5,000  & 21,181  & 132 \\
\midrule
\chemmechqa-Hard
& OOD Test & 8,008 & 2,009 & 6,361 & 11 \\
\bottomrule
\end{tabular}
\end{table}

 \begin{table*}[ht]
\caption{Overview of mechanism and reaction types in \chemmechqa.}
\centering
\begin{tabular}{l p{0.7\textwidth}}
\toprule
\textbf{Category} & \textbf{Space} \\
\midrule
Mechanism Type & Pericyclic, Rearrangement, Radical, Electron transfer, Substitution, Cleavage, Addition, Elimination, Coordination, Proton transfer / Dissociation, Organometallic, Activation, Condensation \\
Mechanism Subtype & Deprotonation, Proton transfer, Elimination, Neutralization of protonated ether / Addition of alcohol under basic pKa, Nucleophilic Addition, Proton exchange, Amine addition, Halide leaves, Alkoxide leaves, Hydroxide addition, ...\\
Hard Reaction Type & Appel bromination (index 5), Carboxylic acid + sulfonamide condensation (index 6), Diels-Alder cycloaddition (index 9), Hydroxy to bromo (index 1), Menshutkin reaction (index 8), N-Bn deprotection (index 11), Negishi coupling (index 2), Prilezhaev epoxidation (index 10), Staudinger reduction (index 4), Transamidation (index 3), Urea Schotten-Baumann (index 7)\\
\bottomrule
\end{tabular}%
\label{tab:CMKG_space}
\end{table*}




\section{Implementation Details and Experiment Settings}\label{app:implementation}
\noindent\textbf{Foundation LLMs.}
We compare GPT-5.2~\cite{singh2025openai}, Gemini-3.1-Pro~\cite{team2023gemini}, DeepSeek-V4-Pro~\cite{deepseekai2026deepseekv4}, Llama-3.1-8B~\cite{grattafiori2024llama}, Qwen2.5-7B~\cite{hui2024qwen2}, and ChemDFM~\cite{zhao2025developing}. 

\noindent\textbf{Mechanistic Models.}
We evaluate three expert models as baselines for chemical mechanistic reasoning, including Reactron~\cite{chen2025predicting}, FlowER~\cite{joung2025electron}, and Graph2SMILES~\cite{joung2024reproducing}.
\begin{itemize}
    
    \item \textbf{Reactron} models electron movement to predict reaction products and arrow-pushing mechanisms.
    \item \textbf{FlowER} is a flow-matching-based generative model designed to model electronic rearrangement pathways in a continuous latent space.
    We adopt the publicly released implementation of this model and retain its default network architecture.
    We retrain the model on \chemmechqa. 
    Due to the scale difference between datasets, we slightly adjust a subset of training hyperparameters.
    The modified training hyperparameters are: warmup\_steps = 3,000, epoch = 200, max\_steps = 3,000,000, save\_iter = 3,000, log\_iter = 10, and eval\_iter = 3,000.
    When retraining and evaluating the FlowER model, we use atom-mapped versions of \chemmechqa, as required by the model’s architecture. 
    The same applies to Graph2SMILES.
    The model is trained for elementary reaction prediction.

    \item \textbf{Graph2SMILES} is a graph-to-sequence model that encodes molecular graphs using DGCN and decodes reaction outcomes as SMILES sequences with a Transformer.
    The reproduction method of Graph2SMILES is similar to that of FlowER. 
    The modified training hyperparameters are: warmup\_steps = 800, epoch = 2,000, log\_iter = 10, eval\_iter = 500, and save\_iter = 1,000.
\end{itemize}

\noindent \textbf{GCN-based mechanism prediction.} 
The model architecture for mechanism prediction follows an encoder--decoder for link prediction. Specifically, the encoder is a two-layer GCN, and the decoder is an MLP.
For training, existing relations in the dataset are treated as positive edges. 
For each positive edge, a template node is randomly sampled for its corresponding molecule node to construct a negative edge. 
This strategy preserves the same bipartite node-type constraints as the positive edges while generating non-existent molecule–template connections. 
Each positive edge is paired with two negative edges, resulting in a balanced 1:2 ratio between positive and negative samples.
The encoder and decoder are optimized jointly with a binary link prediction loss. 
We train GCN-based predictors for 1000 epochs using Adam with a learning rate of $1\times10^{-3}$. The dropout rate is set to 0.5.

\textbf{Metrics.} Previous works~\cite{joung2024reproducing, joung2025electron} primarily evaluate model performance at the level of single-step elementary reactions. In this work, we focus on the more challenging task of multi-step mechanism imputation. To comprehensively assess pathway prediction quality, we adopt five evaluation metrics: Validity, Main Product Accuracy (MPA), Intermediate Hit (I-hit), Pathway Accuracy (PA), and Sequence Rank (SR). Specifically, the \textbf{Validity} metric measures the proportion of chemically valid molecules generated among the predicted mechanistic pathways. The \textbf{Main Product Accuracy (MPA)} metric evaluates whether the predicted main product exactly matches the ground-truth main product. The \textbf{Intermediate Hit (I-hit)} measures intermediate recall by calculating the proportion of correctly predicted intermediates among all ground-truth intermediates across the dataset. \textbf{Pathway Accuracy (PA)} measures the proportion of samples whose entire predicted molecular mechanistic pathways exactly match the reference pathways, thereby evaluating end-to-end pathway reconstruction capability. Additionally,  \textbf{Sequence Rank (SR)} measures whether the top-1 predicted mechanistic pathway matches any potential pathway within the \chemmechkg. 
For pathway prediction, each reported score is the arithmetic mean of the corresponding metric over five runs. For predictions without Gemini refinement, top-$k$ MPA and PA use cumulative hits among the first $k$ candidate pathways. For Gemini-based variants, a separate candidate set is generated for each value of $k$, and MPA and PA measure cumulative hits within that set. These sets are not required to be nested across $k$, so the reported scores need not increase with $k$. Top-$k$ Validity and I-hit are arithmetic means of the corresponding candidate-level scores.

\definecolor{promptaccent}{HTML}{0072B2}
\newcommand{\promptfield}[1]{%
    {\normalfont\sffamily\bfseries\fontsize{8pt}{10pt}\selectfont\color{black}#1}%
}
\newcommand{\promptslot}[1]{\textcolor{promptaccent}{\textit{#1}}}
\newcommand{\promptsection}[1]{%
    \par\vspace{2pt}%
    {\color{gray!40}\hrule height 0.3pt}%
    \vspace{2pt}\noindent\promptfield{#1}\par\nobreak
}
\newcommand{\promptchoice}[1]{%
    \par\noindent\hangindent=1.5em\hangafter=1\textbf{#1:}\enspace
}

\newenvironment{appendixprompt}[3][Prompt]{%
    \par\begingroup\centering
    \def\savedpromptcaption{#2}\def\savedpromptlabel{#3}%
    \captionsetup{position=top}%
    \edef\emitpromptcaption{\noexpand\captionof{table}{\noexpand\textcolor{black}{\savedpromptcaption}}\noexpand\label{\savedpromptlabel}}%
    \emitpromptcaption
    \vspace{-6pt}
    \fontsize{8pt}{9.2pt}\selectfont
    \setlength{\tabcolsep}{5pt}%
    \begin{tabular}{|p{\dimexpr\linewidth-2\tabcolsep-2\arrayrulewidth\relax}|}%
    \hline
    \rowcolor{gray!12}\promptfield{#1}\\
    \hline
    \begin{minipage}{\linewidth}
    \raggedright
    \setlength{\parindent}{0pt}%
    \setlength{\parskip}{2pt}%
    \vspace{3pt}\ignorespaces
}{%
    \par\vspace{3pt}%
    \end{minipage}\\
    \hline
    \end{tabular}
    \vspace{8pt}
    \par\endgroup
}

\clearpage
\begin{figure*}[!p]
\section{Cases}\label{app:cases}
Figures~\ref{fig:case_study1_2} and~\ref{fig:case_study3} show qualitative judgments of predicted and reference pathways, rather than experimental validation.
    \centering
    \includegraphics[width=\textwidth,height=0.85\textheight,keepaspectratio]{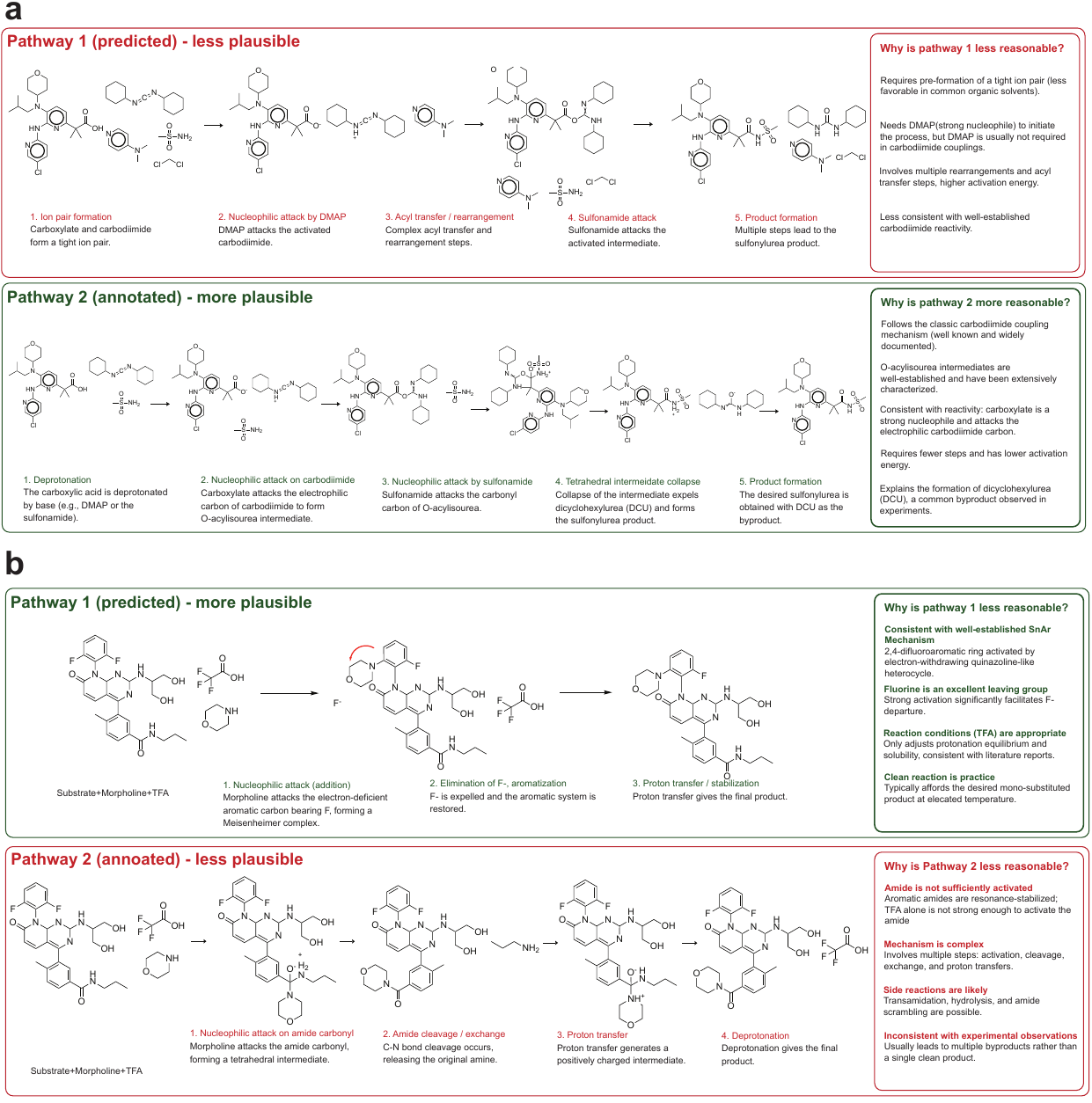}
    \caption{
    Qualitative pathway comparisons: \textbf{a}, a failure case; \textbf{b}, a prediction judged more reasonable than the reference by experts.
    }
    \label{fig:case_study1_2}
\end{figure*}

\begin{table*}[!p]
\begin{minipage}[t]{0.48\textwidth}
\vspace{0pt}
\begin{appendixprompt}[Mechanism pathway prediction with retrieval-guided refinement]{Prompt for GRMP in Mechanism Pathway Prediction with Retrieval Refinement}{tab:prompt_retrieval}

\promptfield{Question:} Please generate the product SMILES based on the given reactant SMILES.

\promptfield{\# Topic entities:} \promptslot{[input SMILES]}

{Use the predicted molecular-state pathway and retrieved graph triplets below to assess and refine the prediction.}

\promptfield{Predicted pathway:} \promptslot{[input predicted pathway]}


\promptfield{Retrieved graph triplets:} \promptslot{[input retrieved triplets]}

{Check SMILES validity and chemical consistency using the retrieved triplets. Correct the predicted pathway if needed; otherwise retain it. Output only the ordered molecular-state SMILES in a code block, separated by arrows. Do not output template SMARTS or explanatory text. Do not leave the output empty.}
\end{appendixprompt}

\end{minipage}\hfill
\begin{minipage}[t]{0.48\textwidth}
\vspace{0pt}
\begin{appendixprompt}[LLM-as-Judge: evaluation rubric]{Rubrics and Prompt for LLM-as-Judge}{tab:prompt_judge}

{Assess the chemical plausibility of the supplied molecular-state pathway using the reactants, reaction conditions, and rubric below.}

{\promptfield{Task:} Rate the supplied predicted pathway.}

\promptfield{\# Topic entities:} \promptslot{[input SMILES]}

{Evaluate the supplied pathway using these expert-defined criteria:}

\promptfield{Predicted pathway:} \promptslot{[input predicted pathway]}

\promptsection{Rubrics:}
\begin{enumerate}[leftmargin=1.5em,labelsep=0.4em,itemsep=1pt,topsep=2pt,parsep=0pt]
    \item \textbf{Intermediate Structural Validity:} Intermediate molecular states should have chemically valid structural representations, including appropriate valence, bond orders, and formal charges. Do not penalize justified reactive, charged, or radical intermediates merely for being unstable. Assess intermediate states here; assess the final product under Product Electronic Validity.
    
    \item \textbf{Mechanistic Plausibility:} All mechanistic steps and the predicted final product must be chemically consistent with the provided reactants and reaction conditions;
    
    \item \textbf{Electron-Transfer Coherence:} The reasoning process must avoid self-reinforcing or circular electron-transfer loops;
    
    \item \textbf{Pathway Completeness:} The pathway should include the essential stages needed to complete the proposed transformation, without premature termination or omission of necessary steps. 
    
    \item \textbf{Product Electronic Validity:} The final predicted product should not contain unpaired or artificially retained electrons.

\end{enumerate}

{Assign each criterion a numeric score between 0.0 and 1.0, inclusive, with higher scores indicating better satisfaction of the criterion. Provide a brief rationale for each score.
}

\end{appendixprompt}
\end{minipage}
\par\medskip
    \centering
    \includegraphics[width=\textwidth,height=0.27\textheight,keepaspectratio]{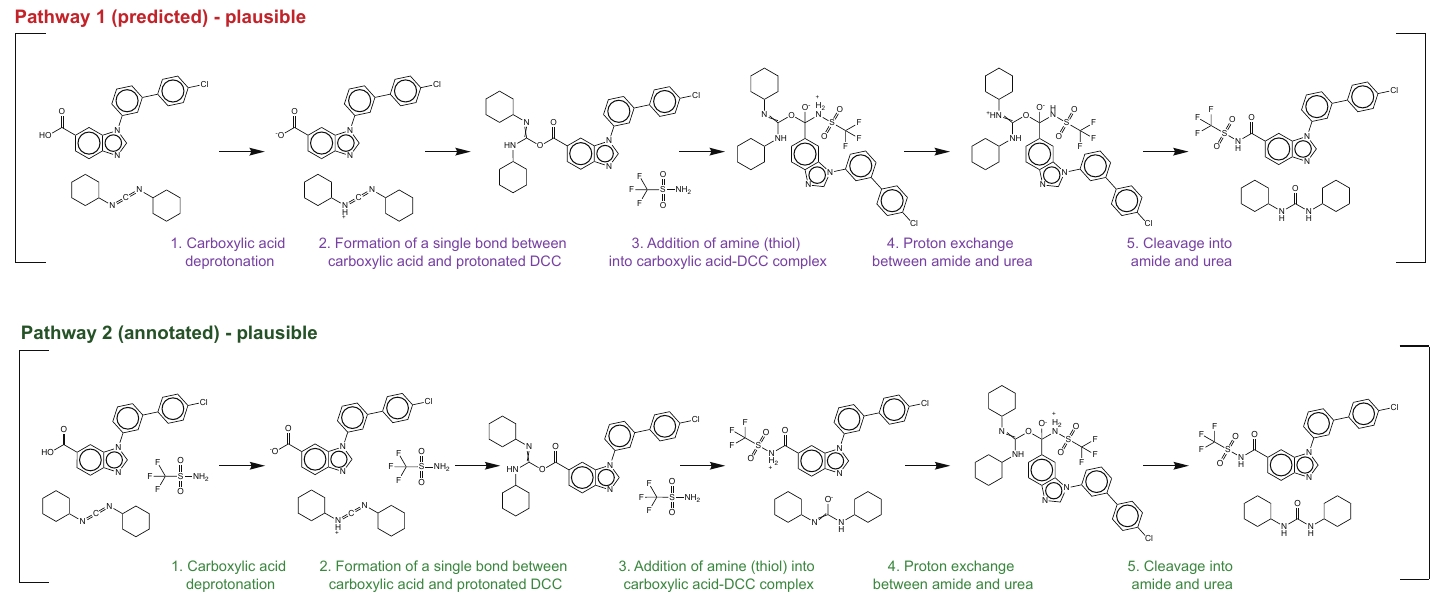}
    \captionof{figure}{
    The LLM judge accepts both pathways; human evaluators prefer the prediction. Neither route is experimentally validated by these judgments.
    }
    \label{fig:case_study3}

\end{table*}

\end{document}